\documentclass[a4paper, conference]{IEEEtran}
\usepackage{cite}
\usepackage{amsmath,amssymb,amsfonts}
\usepackage{algorithmic}
\usepackage{graphicx}
\usepackage{textcomp}
\usepackage{xcolor}
\usepackage{amsmath}
\usepackage{amssymb}
\usepackage{cite}
\usepackage{amsmath,amssymb,amsfonts}
\usepackage{algorithmic}
\usepackage{graphicx}
\usepackage{textcomp}
\usepackage{xcolor}
\usepackage{amsmath}
\usepackage{amssymb}

\makeatletter
 \let\old@ps@headings\ps@headings
 \let\old@ps@IEEEtitlepagestyle\ps@IEEEtitlepagestyle
 \def\confheader#1{%
 \def\ps@headings{%
 \old@ps@headings%
 \def\@oddhead{\strut\hfill#1\hfill\strut}%
 \def\@evenhead{\strut\hfill#1\hfill\strut}%
 }%
 \def\ps@IEEEtitlepagestyle{%
 \old@ps@IEEEtitlepagestyle%
 \def\@oddhead{\strut\hfill#1\hfill\strut}%
 \def\@evenhead{\strut\hfill#1\hfill\strut}%
 }%
 \ps@headings%
 }
 \makeatother

\def\BibTeX{{\rm B\kern-.05em{\sc i\kern-.025em b}\kern-.08em
    T\kern-.1667em\lower.7ex\hbox{E}\kern-.125emX}}
 \author{\IEEEauthorblockN{Om Khare}
    \IEEEauthorblockA{\emph{Dept. of Computer Engineering and IT}\\
\emph{COEP Technological University}\\
                      Pune, India\\
                      khareom20.comp@coeptech.ac.in}
    \and
    \IEEEauthorblockN{Shubham Gandhi}
    \IEEEauthorblockA{\emph{Dept. of Computer Engineering and IT}\\
\emph{COEP Technological University}\\
                      Pune, India\\
                      shubhammg20.comp@coeptech.ac.in}
    \and
    \IEEEauthorblockN{Aditya Rahalkar}
    \IEEEauthorblockA{\emph{Dept. of Mechanical Engineering}\\
\emph{COEP Technological University}\\
                      Pune, India\\
                      rahalkaram20.mech@coeptech.ac.in}
    \and
    \IEEEauthorblockN{Sunil Mane}
    \IEEEauthorblockA{\emph{Dept. of Computer Engineering and IT}\\
\emph{COEP Technological University}\\
                      Pune, India\\
                      sunilbmane.comp@coeptech.ac.in}
}
\begin{document}

\title{YOLOv8-Based Visual Detection of Road Hazards: Potholes, Sewer Covers, and Manholes}

\maketitle 

\vspace{-5pt}

\begin{abstract}

Effective detection of road hazards plays a pivotal role in road infrastructure maintenance and ensuring road safety. This research paper provides a comprehensive evaluation of YOLOv8, an object detection model, in the context of detecting road hazards such as potholes, Sewer Covers, and Man Holes. A comparative analysis with previous iterations, YOLOv5 and YOLOv7, is conducted, emphasizing the importance of computational efficiency in various applications. The paper delves into the architecture of YOLOv8 and explores image preprocessing techniques aimed at enhancing detection accuracy across diverse conditions, including variations in lighting, road types, hazard sizes, and types. Furthermore, hyperparameter tuning experiments are performed to optimize model performance through adjustments in learning rates, batch sizes, anchor box sizes, and augmentation strategies. Model evaluation is based on Mean Average Precision (mAP), a widely accepted metric for object detection performance. The research assesses the robustness and generalization capabilities of the models through mAP scores calculated across the diverse test scenarios, underlining the significance of YOLOv8 in road hazard detection and infrastructure maintenance.

\end{abstract}

\begin{IEEEkeywords}
YOLO, Object Detection, Autonomous, Deep learning, Potholes, Sewer Covers, Man Holes
\end{IEEEkeywords}

\section{Introduction}

In the era of advancing transportation systems, the goal of enhancing transportation infrastructure for safety and efficiency has come to the forefront. Factors such as heavy rainfall, inadequate road maintenance, and the possibility of natural disasters have highlighted the urgent need for swift and efficient road hazard detection. Potholes, Sewer Covers, and Man Holes pose substantial threats to both motorists and pedestrians, leading to approximately 4,800 \cite{e1} accidents annually and placing a significant financial burden on public resources due to vehicle damages and related expenses. The rugged contours of potholes have the potential to cause a range of damages, from tire tears to compromising wheel rim integrity upon impact, presenting risks that could potentially lead to fatal consequences.

One notable innovation in the field of Computer Vision is the You Only Look Once version 8 (YOLOv8) object detection algorithm developed by Ultralytics et al. YOLOv8 is a state-of-the-art deep learning model that has demonstrated success in various real-time object detection applications. This paper investigates the potential effectiveness of YOLOv8 as a robust tool for Pothole, Sewer Cover, and Man Hole detection. It conducts a comprehensive analysis and comparative evaluation, shedding light on its performance compared to its predecessors, YOLOv7 and YOLOv5.

\subsection{Problem Statement}
Traditional methods of pothole detection often rely on manual inspections or sensor-based systems, both of which exhibit limitations in terms of accuracy, scalability, and real-time responsiveness. Manual inspections are time-consuming, subjective, and impractical for covering extensive road networks efficiently. On the other hand, sensor-based systems can be expensive to deploy and maintain, and their effectiveness can be hindered by various environmental factors. Furthermore, many existing computer vision approaches for pothole detection lack the speed and accuracy required for real-time implementation, especially in the context of autonomous vehicles.

While there have been previous attempts to use deep learning models like YOLOv7 and YOLOv5 for pothole detection, these models have their own set of disadvantages. YOLOv7, for example, may exhibit lower accuracy and efficiency compared to newer versions, while YOLOv5 might not offer the same level of robustness required for diverse environmental conditions. Addressing these challenges is crucial for enhancing road safety, minimizing vehicle damage, and optimizing road maintenance efforts.

\begin{table*}[h]
\caption{Approaches used in Pothole detection}
\label{table:approaches}
\centering
\renewcommand{\arraystretch}{1.2}
\begin{tabular}{|p{2cm}|p{4cm}|p{2cm}|p{4cm}|p{4cm}|}
\hline
\textbf{Methodology} & \textbf{Abstract} & \textbf{Reference} & \textbf{Pros} & \textbf{Cons} \\
\hline
Vibration based & Uses GPS, accelerometer, gyroscope units for mapping road surfaces. Wavelet decomposition \& SVM used & [2], [3], [4] & Real-time insights into road network conditions. Approx. 90\% accuracy in severe anomalies. & Requires vehicles to drive over potholes. Limited to devices with specific hardware/software capabilities. \\
\hline
3D Laser-based & Uses 3D laser scanning to identify pavement distresses. Grid-based approach for specific distress features. & [5], [6] & Accurate 3D point-cloud points. FNN for severity classification. & Costly, short range of detection. Not suitable for early detection by autonomous vehicles. \\
\hline
3D Stereo-vision & Reconstructs 3D pavement surface from input images. Uses stereo images for road distress identification. & [7], [8] & Precise representation of road surfaces, Hence high accuracy. & Requires high computational power for 3D surface reconstruction. \\
\hline
Vision-based (2D) & Uses CNN, DNN for road damage detection using a dataset of road damage images for training. & [9], [10], [11], [12] & Cost-effective, enables determining the exact shape and area of a pothole. High accuracy. & There is a tradeoff between dataset diversity, model's accuracy, processing time and model size. \\
\hline
\end{tabular}
\end{table*}

\subsection{Motivation and Objectives}

Motivated by the pressing need for accurate and timely road hazards, the observed advantages of YOLOv8, including increased speed and accuracy, as well as its improved model architecture and user-friendliness, are sought to be evaluated by this study. Firstly, crucial insights into the model's architecture that enable it to excel in real-time identification, its precision, and its lightweight characteristics in identifying road hazards of varying sizes and categories will be offered through this evaluation. Secondly, rigorous performance tests and benchmarking are conducted to compare YOLOv8 with its predecessors, namely YOLOv7 and YOLOv5, and ascertain how its enhancements in speed and accuracy surpass those of the earlier models.

\section{Related Work}

In recent years, there has been a significant surge in research focusing on road conditions, encompassing challenges like potholes, manholes, sewer covers, and man hole detection. This heightened interest can largely be attributed to the advancements in autonomous vehicle technologies, where the accurate mapping of road conditions holds paramount importance. Pothole detection methods have evolved into various categories,\cite{1} including vibration-based, 3D laser-based, 3D reconstruction, and 2D vision-based approaches. Table \ref{table:approaches}, outlines the strengths and limitations associated with each of these approaches.

In vibration-based methods, hazards are detected using accelerometers. A vibration-based system was developed to estimate pavement conditions \cite{2}. It models the interactions between the ground and the vehicle, considering the vehicle to be under random force excitations. Real-time detection of road irregularities, potential hazards, is achieved using a mobile sensing system that utilizes the accelerometers in smartphones\cite{3}. These devices were specifically designed for limited access to hardware and software and did not require extensive signal-processing techniques.

\cite{4} explored the use of devices equipped with GPS, accelerometers, and gyroscope units to map road surfaces. Wavelet decomposition was employed for signal processing, while Support Vector Machine (SVM) was utilized for the detection of cracks and irregularities on the road surfaces. They consistently achieved an accuracy of approximately 90\% in detecting severe anomalies, regardless of the vehicle type or road location, providing real-time insights into road network conditions.

3D construction methods are further classified into laser-based and stereo-vision approaches. The potential of 3D laser scanning as a tool for identifying pavement distresses, such as potholes, was explored \cite{5}. The 3D laser scanning technology captured accurate 3D point cloud data, which was then processed using a grid-based approach to focus on specific distress features. \cite{6} utilized laser imaging to find distress in pavements. This method captured pavement images and identified pothole areas, which were then represented using a matrix of square tiles. A feedforward neural network (FNN) was used to classify the severity of pothole and crack types, demonstrating its capability to enhance pavement images, extract potholes, and analyze their severity. \cite{7} introduced a novel approach using the stereovision technique to reconstruct a full 3D pavement surface from the input images. The methodology involved calibrating the input data, correcting any observed distortion, feature extraction, and 3D reconstruction. \cite{8} introduced a method that created a set of points in a three-dimensional space, allowing for a precise representation of road surfaces. By leveraging stereo images and image processing technologies, the system could identify various road distresses, such as potholes, bumps, and cracks, etc
\begin{figure*}[ht]
  \centering
  \includegraphics[width=0.8\textwidth]{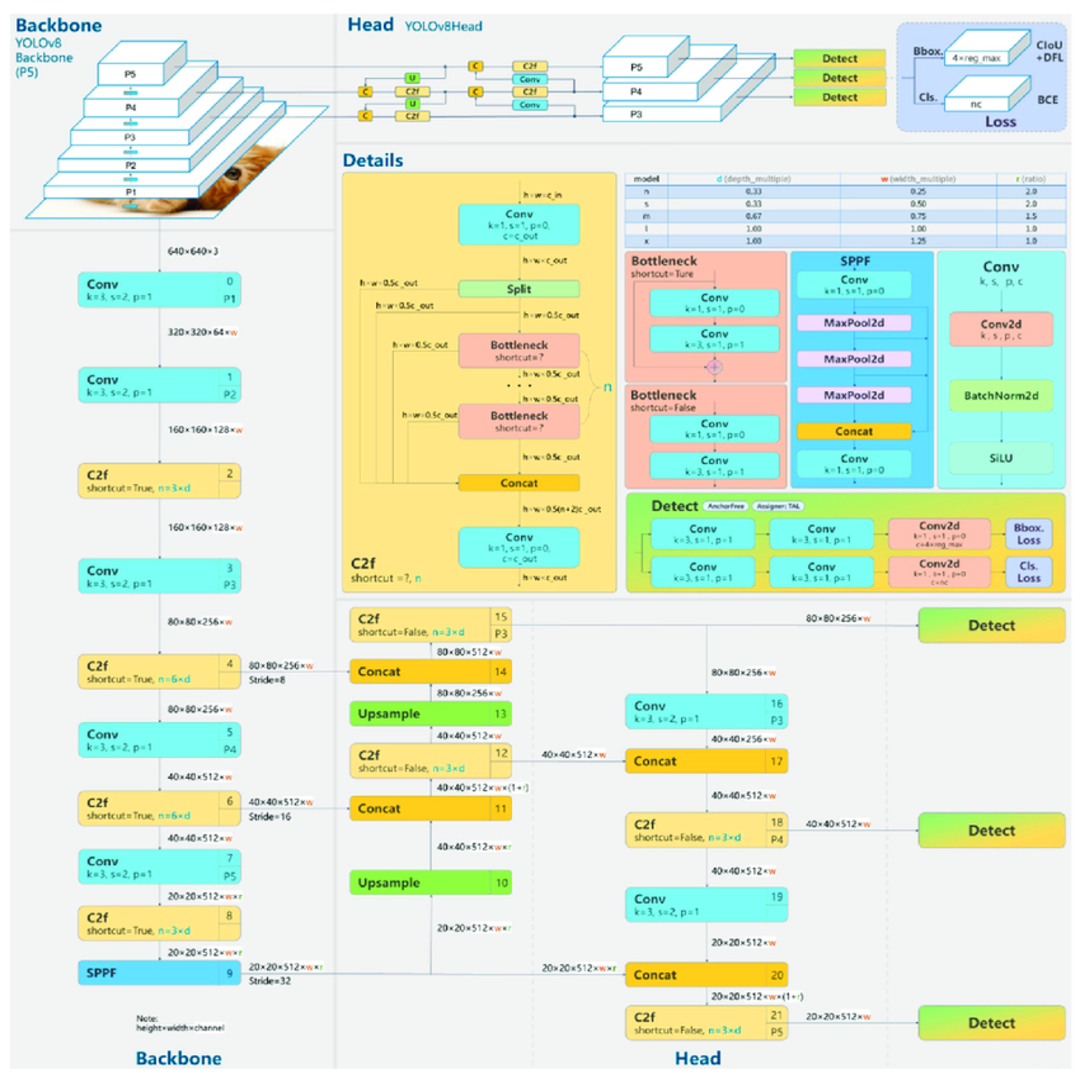}
  \caption{YOLOv8 Architecture}
  \label{fig:arc}
\end{figure*}

Vision-based methods utilize image processing and deep learning on 2D images obtained from cameras. A system that uses Convolutional Neural Networks (CNNs) was proposed to detect road damage \cite{9}. A large dataset comprising irregularities on road images captured via smartphones, with several instances of road surface damage, was used, achieving an accuracy of 75\%. \cite{10} used thermal images as input feed for deep neural networks for the detection and local mapping of potholes. To deal with the challenges posed by changing weather conditions, a modified ResNet50-RetinaNet model was employed, achieving a precision of 91.15\% in pothole localization using thermal images. \cite{11} explored the use of YOLO (You Only Look Once), a model that uses a combination of region proposal algorithms and CNNs to detect and localize objects in images. They developed a new dataset comprising 1500 images of Indian roads with 76\% as the highest precision obtained from YOLOv3 Tiny. \cite{12} used YOLOv7, leveraging the power of Convolutional Neural Networks for detecting potholes. An open-source dataset was used for training the model, which achieved an F1 score of 0.51.

The following drawbacks have been identified in current implementations:

\begin{enumerate}
    \item Insufficient diversity in the datasets utilized for training and testing the model.
    \item High processing time.
    \item Large model size.
    \item Low mAP@0.5 metric values for the deployed model.
\end{enumerate}

\section{Methodology}

In this section, we discuss the architecture, dataset, preprocessing methods, and hyperparameter tuning used in training the YOLOv8 model.

\subsection{Architecture}
The architecture of YOLOv8, illustrated in Figure \ref{fig:arc}\cite{arcdiag}, incorporates novel improvements for superior detection accuracy while maintaining high speed and efficiency. YOLOv8's backbone architecture is inspired by YOLOv5 but includes several key modifications \cite{arc1}:

\begin{enumerate}
    \item \textbf{CSPDarknet53 Feature Extractor}: YOLOv8 employs CSPDarknet53, a variant of the Darknet architecture, as its feature extractor. This component consists of convolutional layers, batch normalization, and SiLU activation functions. Notably, YOLOv8 replaces the initial 6x6 convolutional layer with a 3x3 convolutional layer for improved feature extraction.
    
    \item \textbf{C2f Module (Cross-Stage Partial Bottleneck)}: YOLOv8 introduces the C2f module to effectively merge high-level features with contextual information. This is achieved by concatenating the outputs of bottleneck blocks, which consist of two 3x3 convolutions with residual connections. This architectural change aims to enhance feature representation.
    
    \item \textbf{Detection Head}: YOLOv8 adopts an anchor-free detection strategy, eliminating the need for predefined anchor boxes and directly predicting object centers. The detection head comprises the following components:
 
   \begin{enumerate}
            \item \textbf{Independent Branches}: YOLOv8 uses a decoupled head approach, where objectness, classification, and regression tasks are processed independently by separate branches. This design allows each branch to focus on its specific task, contributing to overall detection accuracy.
            
            \item \textbf{Activation Functions}: The objectness score in the output layer uses the sigmoid activation function, representing the probability of an object's presence within a bounding box. For class probabilities, YOLOv8 employs the softmax function, indicating the likelihood of an object belonging to each class.
            
            \item \textbf{Loss Functions}: To optimize the model, YOLOv8 utilizes the CIoU (Complete Intersection over Union) and DFL (Dynamic Focal Loss) loss functions for bounding box regression and binary cross-entropy for classification. These loss functions are effective in enhancing object detection, especially for smaller objects.
        \end{enumerate}
    
    \item \textbf{YOLOv8-Seg Model}: In addition to object detection, YOLOv8 offers a semantic segmentation model known as YOLOv8-Seg. This model uses CSPDarknet53 as its backbone feature extractor and incorporates the C2f module. Two segmentation heads are added to predict semantic segmentation masks, making it a versatile tool for various computer vision tasks.
\end{enumerate}

\subsection{Training Techniques}

To enhance model performance, YOLOv8 employs innovative training techniques, including Mosaic Augmentation. During training, YOLOv8 combines four images, encouraging the model to learn object contexts in different locations and against varying backgrounds. However, this augmentation is disabled during the final ten training epochs to prevent potential performance degradation.

\subsection{Dataset}

The dataset used for pothole detection is sourced from Roboflow Universe, specifically the "Pothole Detection" dataset provided by the Intel Unnati Training Program \cite{13}. This dataset offers a comprehensive collection of annotated images, classifying not only potholes but also manholes and sewer covers. This diversity aids in making the model more robust and preventing misclassification. The dataset comprises images captured under various lighting conditions, angles, and environments.

The Indian Driving Dataset is included in this dataset, enabling the model to train on images captured from a dashboard camera's perspective. Additionally, the dataset contains images of water-filled potholes, ensuring that the model is trained to recognize a variety of scenarios, thus enhancing its robustness. The dataset consists of a total of 3,770 images, with 3,033 images in the training set, 491 images in the validation set, and 246 images in the testing set.

\subsection{Preprocessing}

To enhance the robustness of our model, the dataset underwent several preprocessing methods. Since all input images in the dataset were of size 640 x 640 pixels, no image resizing was required. Min-max normalization was applied to normalize the pixel values across all images.

Data augmentation was performed to increase the dataset's size and diversity, which proved beneficial given the uncertainty in weather conditions, camera mounting, and image quality variability. Five types of modifications were applied:

\begin{enumerate}
    \item \textbf{Image Flip}: The image is horizontally flipped to account for different shapes and orientations of potholes.
    \item \textbf{Image Scaling}: Image scaling helps the model adapt to different pothole sizes.
    \item \textbf{Motion Blurring}: To train the model for low-quality images and images with motion blur, a blur effect was introduced.
    \item \textbf{Color Manipulation}: Color manipulation or RGB manipulation helps the model adapt to varying lighting conditions, such as bright sunlight or low ambient light at night.
    \item \textbf{Fog Addition}: Adding fog to images makes the model robust against foggy conditions.
\end{enumerate}

These data augmentations were applied to every alternate image, resulting in five derivative images from the original image. This increased the size of the training set to 10,613 images.

\subsection{Model Training}

Hyperparameter tuning was conducted by considering different parameters and evaluating which ones yielded the best results. All three models, namely, YOLOv5 tiny, YOLOv5 small, YOLOv7, YOLOv8 tiny, and YOLOv8 small, were trained with the same hyperparameters, as shown in Table \ref{table:hyperparameter}, to ensure comparable results. Training of all models was performed on an NVIDIA GeForce RTX 3090 GPU with 10,496 CUDA cores.

\renewcommand{\arraystretch}{1.5}
\begin{table}[h]
\centering
\begin{tabular}{|l@{\hspace{1cm}}|l|}
\hline
\textbf{Hyperparameter} & \textbf{Value} \\
\hline
Epochs & 250 \\
Batch Size & 16 \\
Learning Rate & 0.01 \\
Weight Decay & 0.0005 \\
Optimizer & Adam \\
Momentum & 0.937 \\
\hline
\end{tabular}
\renewcommand{\arraystretch}{1}
\vspace{5pt}
\caption{Training Hyperparameters}
\label{table:hyperparameter}
\vspace{-15pt}
\end{table}

\begin{figure*}[ht]
  \centering
  \includegraphics[width=1.4\columnwidth]{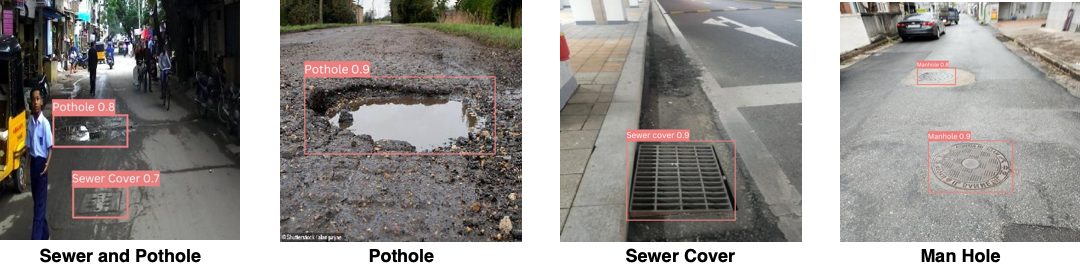}
  \caption{Prediction images by YOLOv8 tiny}
  \label{fig:dsam}
\end{figure*}

\vspace{-5pt}

\section{RESULTS AND DISCUSSION}

\subsection{Performance Metrics}

The following metrics \cite{acc1} were used for evaluating and comparing the models:

\subsubsection{Mean Average Precision}

Mean Average Precision (mAP), computed using equation \ref{eq:eq1}, is a widely accepted performance metric for object detection models. mAP is calculated by taking the mean of Average Precision (AP) for each of the 'n' classes. AP for each class 'k' is determined by calculating the area under the precision-recall curve. mAP provides a single score that considers Recall, Precision, and Intersection over Union (IoU), eliminating bias in performance measurement.

\begin{equation*}
mAP = \frac{1}{n}\sum\limits_{k=1}^{k=n}AP_{k}
\label{eq:eq1}\tag{1}
\end{equation*}

\subsubsection{Processing Time}

Processing time is a crucial metric for assessing the speed at which the model classifies an input image. It encompasses the total time taken by the model for pre-processing, inference, loss calculation, and post-processing of an image. Swift decision-making is essential for autonomous vehicles to avoid potholes, making the minimization of processing time imperative.

\subsubsection{Size of the Trained Model}

The size of the deployed model, responsible for processing incoming data in embedded systems, depends on the size of the final trained model generated. Keeping this size as small as possible is essential due to limitations in onboard memory and the storage capacity of the hardware. Larger models require more computation, resulting in lower power efficiency.

\subsection{Results}

Table \ref{table:accuracy} presents the results obtained for Processing Time and the size of the model on various YOLO models.

\begin{table}[h]
\centering
\begin{tabular}{|l|l|l|l|}
\hline
\textbf{Model} & \textbf{mAP@0.5} & \textbf{Processing Time} & \textbf{Size of Model} \\
\hline
YOLOv5 nano & 0.84 & 28 ms & 14.8 MB \\
YOLOv5 small & 0.86 & 38 ms & 15.1 MB \\
YOLOv7 & 0.90 & 35 ms & 74.8 MB \\
YOLOv8 nano & 0.911 & 8.8 ms & 6.3 MB \\
YOLOv8 small & 0.92 & 11 ms & 21.5 MB \\
\hline
\end{tabular}
\vspace{5pt}
\caption{Model Performance Metrics}
\label{table:accuracy}
\end{table}
\begin{table}[h]
\centering
\begin{tabular}{|l|l|l|}
\hline
\textbf{Class} & \textbf{mAP@0.5} & \textbf{mAP@0.95} \\
\hline
Overall & 0.91 & 0.59 \\
Manhole & 0.979 & 0.606 \\
Pothole & 0.871 & 0.648 \\
Sewer Cover & 0.88 & 0.516 \\
\hline
\end{tabular}
\vspace{5pt}
\caption{Class-wise Mean Average Precision (mAP) Metrics for YOLOv8 nano}
\label{table:class}
\end{table}
\begin{figure}[ht]
  \centering
  \includegraphics[width=0.8\columnwidth]{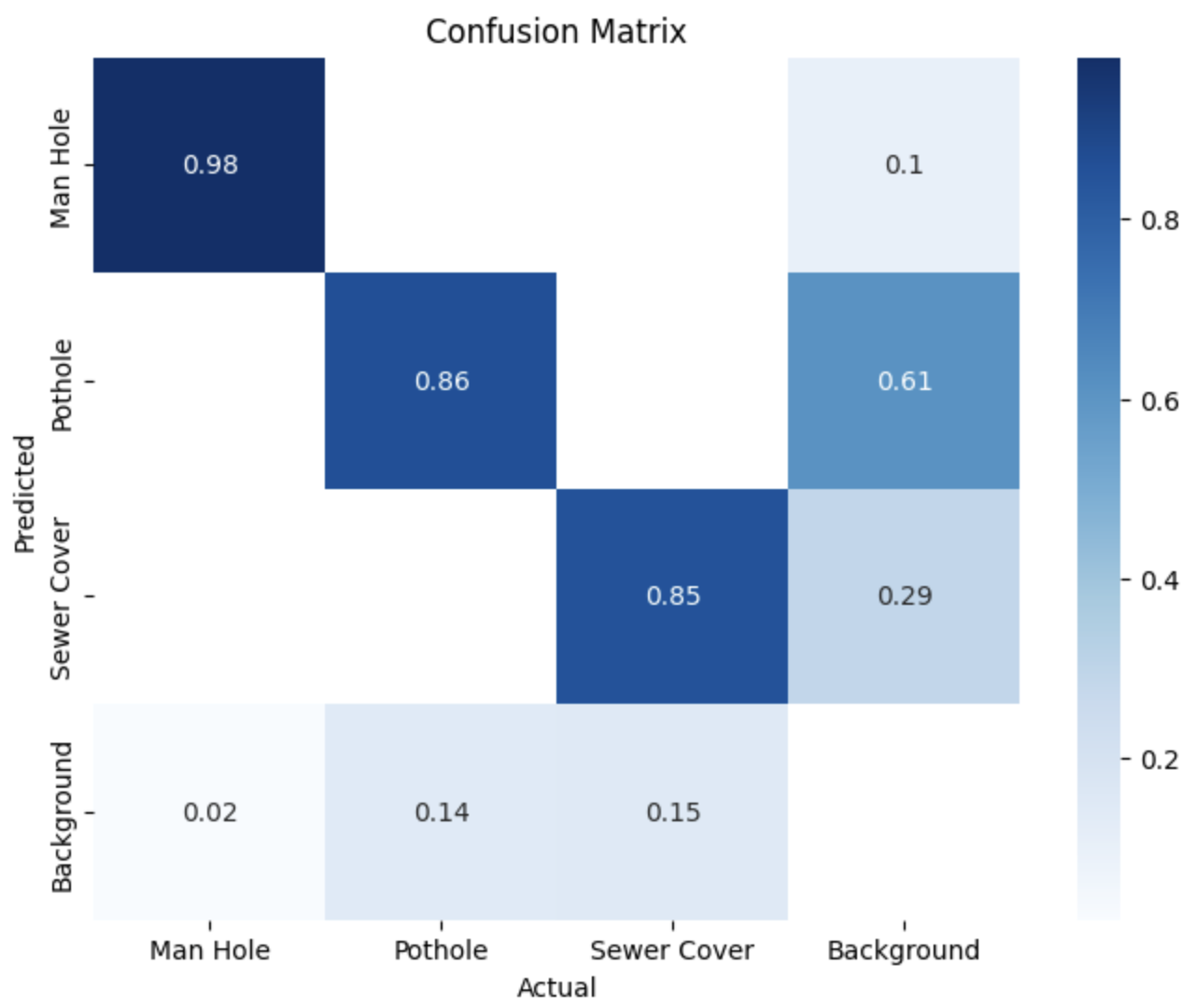}
  \caption{Confusion Matrix}
  \label{fig:cm}
\end{figure}
While YOLOv5 nano and small models have small sizes (14.8 and 15.1 MB, respectively), they yield lower mean average precision compared to other YOLO models. YOLOv7 achieves a good mean average precision value but results in high processing time and model size. YOLOv8 nano and small models outperform the previous YOLO models in every metric. The most efficient model, considering all three parameter metrics, is the YOLOv8 nano model. It provides an exceptional average processing time of 8.8 ms for a single image while maintaining a model size of just 6.3 MB, making it lightweight for deployment in embedded systems.

Table \ref{table:class} shows the results obtained for each class for the YOLOv8 nano model. Our model achieved a precision of 0.90, a recall of 0.81, an mAP of 91\% at 0.5 IoU, and an mAP of 59\% at 0.95 IoU. The results of the YOLOv8 model after applying it to the images in the dataset are presented in Figure \ref{fig:dsam}, and the confusion matrix for the trained model is illustrated in Figure \ref{fig:cm}.

The inclusion of manholes and sewer covers in the training dataset benefits the model by increasing its robustness in correctly identifying potholes, which is crucial for autonomous vehicles to take appropriate action. In the case of manholes and sewer covers, there is no need for the vehicle to take any action, preventing the model from giving false positives. As illustrated in the first image of Figure \ref{fig:dsam}, the model accurately identifies and distinguishes between a sewer cover and a pothole, highlighting the robust nature of our model.

Comparing our implementation of YOLOv8 with previous works based on YOLO, superior performance is displayed across all categories such as mAP, Processing Time, and Size of Model as illustrated in Table \ref{table:compare}.

\begin{table}[h]
\centering
\begin{tabular}{|p{1.8cm}|p{1.1cm}|p{1.5cm}|p{1.3cm}|p{1cm}|}
\hline
\textbf{Reference} & \textbf{Model} & \textbf{mAP (IOU=50\%)} &\textbf{Processing Time} & \textbf{Size of Model} \\
\hline
Ukhwah et. al. \cite{14} & YOLOv3 & 88.93  & 40 ms & N/A \\
Shaghouri et. al. \cite{15} & YOLOv4 & 85.39 & 50 ms & N/A\\
Gao et. al. \cite{16} & YOLOv5 & 93.99 & 12.78 ms & N/A\\
Proposed Implementation & YOLOv8 & 91.11 & 8.8 ms & 6.3 MB\\
\hline
\end{tabular}
\vspace{5pt}
\caption{Comparison of YOLOv8 with previous work}
\label{table:compare}
\end{table}

\section{Conclusion}

In conclusion, this study comprehensively evaluated the performance of the YOLOv8 object detection model in the critical task of pothole detection. The research presented a detailed discussion of YOLOv8's architecture, compared it with earlier iterations, and various versions of YOLOv5 and YOLOv7, including nano and small variants. The evaluation criteria encompassed processing time, model size, and robustness under diverse conditions.

The experiments conclusively establish the superiority of YOLOv8, particularly the nano variant, as the most efficient and effective model for pothole detection. It achieved an impressive mean average precision (mAP) of 0.911 at 0.5 IoU while maintaining a remarkably low processing time of just 8.8 ms per image and a compact model size of 6.3 MB. These attributes are of paramount importance for seamless real-time implementation in the context of road hazard detection, where rapid decision-making and resource efficiency are indispensable.

Furthermore, the research underscores the significance of a diverse and robust training dataset, including not only potholes but also manholes, sewer covers, and various other road elements. This meticulous approach significantly enhanced our model's capability to distinguish between different road hazards, substantially reducing false positives and ensuring the precise detection of potholes.

In summary, the YOLOv8 model, particularly the nano variant, emerges as a highly promising solution for road hazard detection. It strikes a delicate balance between accuracy, speed, and resource efficiency, making significant strides in advancing road safety and infrastructure maintenance. This research not only contributes to the field of road hazard detection but also paves the way for enhanced safety and efficiency in road infrastructure maintenance. Future work may involve real-world deployment and further optimizations to continually enhance its performance in various road and lighting conditions.

\bibliography{references.bib}

\end{document}